\documentclass[a4paper,conference]{IEEEtran}

\ifCLASSINFOpdf
\else
\fi
\usepackage{amsmath}
\usepackage{mathtools}
\usepackage{amsfonts}
\usepackage{xcolor}
\usepackage{textcomp}
\usepackage{multirow}
\usepackage{adjustbox}

\definecolor{orange}{rgb}{0.6, 0.5, 0.0}
\definecolor{cfgreen}{rgb}{0.0, 0.42, 0.24}

\def\etal{\emph{et al.}}
\def\eg{\emph{e.g.}}
\def\ie{\emph{i.e.}}

\begin{document}
%
\title{What makes you, you? \\ Analyzing Recognition by Swapping Face Parts}

\author{\IEEEauthorblockN{Claudio Ferrari}
\IEEEauthorblockA{Department of Architecture and Engineering\\
University of Parma, Italy}
\and
\IEEEauthorblockN{Matteo Serpentoni, Stefano Berretti, Alberto Del Bimbo}
\IEEEauthorblockA{Department of Information Engineering\\
Media Integration and Communication Center (MICC)\\
University of Florence, Italy}}


%


\maketitle

\begin{abstract}
Deep learning advanced face recognition to an unprecedented accuracy. However, understanding how local parts of the face affect the overall recognition performance is still mostly unclear. Among others, \textit{face swap} has been experimented to this end, but just for the entire face. In this paper, we propose to swap facial parts as a way to 
disentangle the recognition relevance of different face parts, like eyes, nose and mouth. In our method, swapping parts from a source face to a target one is performed by fitting a 3D prior, which establishes dense pixels correspondence between parts, while also handling pose differences. Seamless cloning is then used to obtain smooth transitions between the mapped source regions and the shape and skin tone of the target face. We devised an experimental protocol that allowed us to draw some preliminary conclusions when the swapped images are classified by deep networks, indicating a prominence of the eyes and eyebrows region. Code available at https://github.com/clferrari/FacePartsSwap 
\end{abstract}


%
\IEEEpeerreviewmaketitle

\section{Introduction}\label{sec:introduction}
In recent years, thanks to the significant advances in deep neural networks, the task of \textit{face swapping and manipulation} has attracted an increased research interest. In face swapping, the target face is substituted with a source face (different identity than the target face) while maintaining the facial attributes such as expression, skin color, etc. of the target face. This problem is of particular interest because it allows for conducting face analysis under a variety of aspects that could not be possible with manual tools. Given a target face of subject $B$ and a source face of subject $A$, the goal of face swapping is to transfer the appearance of the source face to the target face. After face swapping, the identity of the target face is the same
as that of the source face. Irrespective of the fact that other attributes, such as facial expressions or accessories, \eg, eyeglasses, are transferred or not, a fundamental assumption upon which swapping methods rely on is that the identity of the individual is encoded in his/her entire face. Under this assumption, a constraint that can be enforced to ease the problem is that the identity of the synthesized image must match that of the source image. A common way to force this behavior and improve the image generation quality consists in making use of auxiliary face recognition networks to extract face descriptors, and use this information to enforce an identity-consistent swap~\cite{li2020advancing}. However, this assumption holds only in case the entire face is transferred from the source to the target image.

\begin{figure}[!t]
\centering
\includegraphics[width=0.99\linewidth]{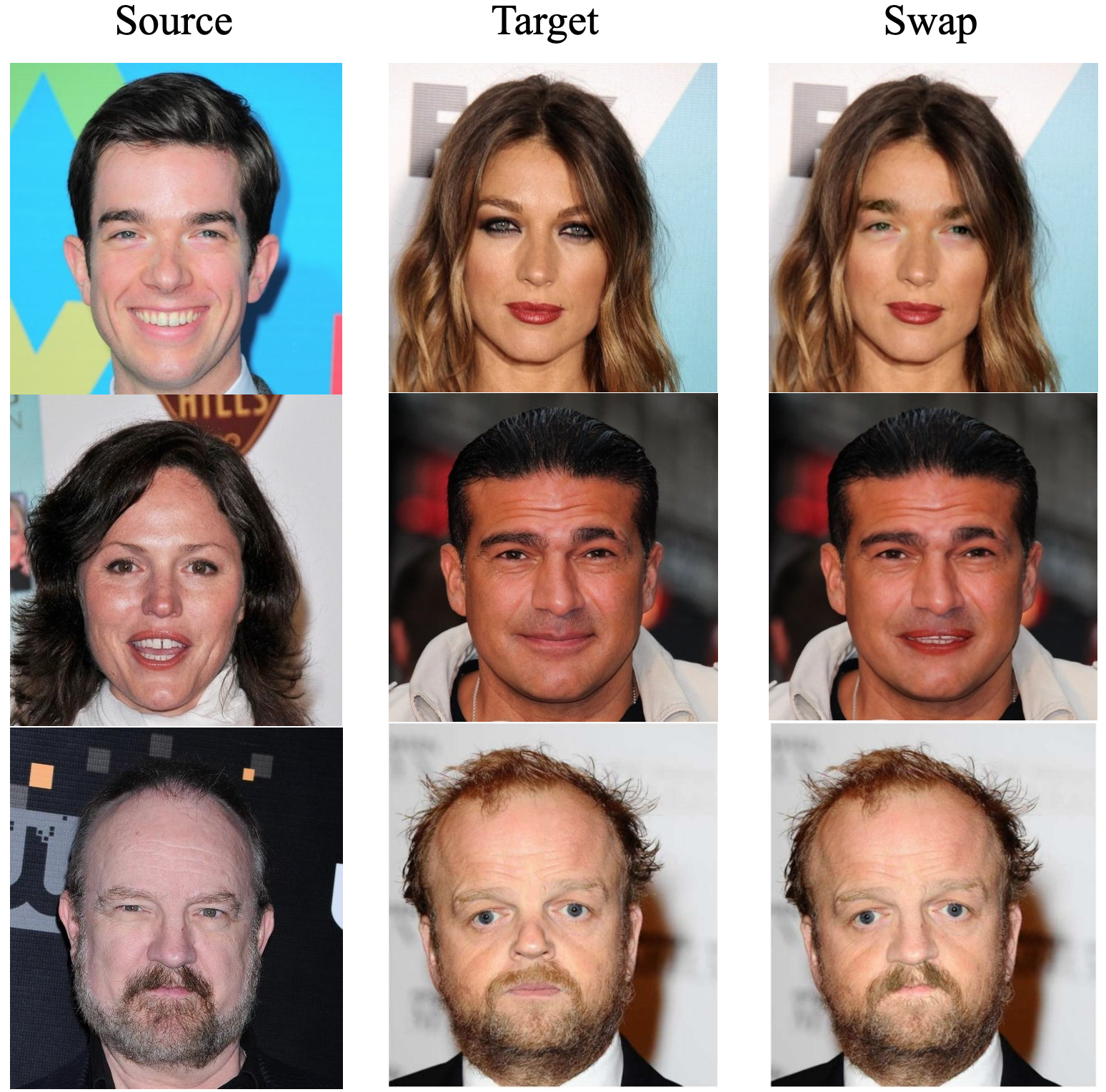}
\caption{Example of local face parts swapping from source to target. The swapped parts are the eyes (top row), the mouth (middle row), and the nose (bottom row).}
\label{fig:eyecatcher}
\end{figure}

In this paper, we propose and investigate a different problem: instead of swapping the entire face, our goal is to transfer local face parts, for example nose or mouth. In this scenario, the above identity-consistency assumption does not hold anymore, and the identity information cannot be used to guide the swapped image synthesis. In fact, when a facial part is changed, we cannot uniquely determine whether the identity is either of the source, the target, or neither of them. So, we have two different scenarios: \textit{(i)} the generated face is a new identity that looks similar to the target subject, or \textit{(ii)} the generated identity is the target subject that changed its appearance in some way, for example after a surgical operation or if in disguise. In both the cases though, the condition is ambiguous. Moreover, depending on the facial part (either one or more) that is swapped, the perceived identity could change more or less significantly. Some swaps examples are shown in Figure~\ref{fig:eyecatcher}: the identity of the swapped image in the top row (swapped eyes), looks more different from the target with respect to, for example, the third row, where the nose was swapped. Clearly, this perceived difference also depends on the prior difference of the source and target faces.

The goal of this work is to extend upon previous research on face perception~\cite{nirkin2018face, axelrod2010external, sinha1996think, ferrari2018investigating} and to study the effects of face swapping and facial parts swapping on perceived identity. The end goal is to evaluate how different facial parts influence the performance of automatic recognition systems based on deep networks.vTo address the problem and overcome the identity ambiguity issue, we establish a baseline method and propose a solution based on a 3D Morphable Model (3DMM) of the face. We first use the 3DMM to fit both the source and target faces, and select the region we wish to swap. Thanks to the fixed topology of the 3DMM, we can put in correspondence triangular image patches defined by the fitted 3DMM, and transfer the image area by applying a piece-wise affine transformation from the source to the target. The final image is obtained by applying a seamless clone algorithm to the swapped image. In summary, the contributions of this work are:
\begin{itemize}
    \item We define a new task known as partial face swapping, and propose a baseline method that works in unconstrained conditions.; 
    \item We perform an investigation of how recognition systems respond to local changes of the face, and analyze the connections with human face perception.
\end{itemize}


\section{Related Work}\label{sec:related}
\noindent
\textbf{Face analysis and perception:} several studies investigated the problem of face recognition and perception, both in humans and machine vision systems, with the goal of understanding how deep networks process face images to collect relevant information for recognition. Some major results from human perception highlighted the following critical aspects: first, identity perception is strongly influenced by context and external cues, such as hair, skin tone or head shape, and not only by internal face features~\cite{sinha1996think, axelrod2010external}. This outcome has been also verified in vision systems. Ferrari~\etal~\cite{ferrari2018investigating} showed that by training deep recognition networks using face images including larger context can significantly improve the recognition accuracy. 
From a different perspective, Nirkin~\etal~\cite{nirkin2018face} showed that, when swapping faces with different external details, strong manipulation is required to correctly match the swapped face with its original counterpart. 

Other interesting aspects of face perception were studied by Sinha~\etal~\cite{sinha2006face}. Among their conclusions, a result is that the image resolution does not play a significant role for recognition, and even low resolution, blurry images lead to good recognition accuracy~\cite{ferrari2018investigating}. On the other hand, it was recently shown how deep networks are quite uninfluenced by the shape of face parts, but strongly biased towards texture~\cite{masi2019face, geirhos2018imagenet}. Guided by this result, Masi~\etal~\cite{masi2019face} reduced the texture bias by changing the underlying face shape, so augmenting the training data with face-specific shape transformations. Here, we go a step further and analyze how different facial features influence recognition. To this aim, we developed a method to swap local face parts, so that both shape, texture and context are mixed up.

\textbf{Face swapping and manipulation:}
\noindent
face swapping and manipulation is a research field that gained a lot of interest recently, thanks to the development of powerful deep generative models. The first swapping methods date back to almost a decade ago~\cite{bitouk2008face, blanz2004exchanging}, and were mainly developed for privacy purposes. Among those, the work by Mosaddegh~\etal~\cite{mosaddegh2014photorealistic} is similar to ours, as they swap and mix different face components from various subjects. However, the method works well only in constrained conditions, where the faces need to be very accurately aligned. To overcome this latter problem and extend to more unconstrained conditions, 3D face models were also employed to account for pose differences~\cite{lin2012face, nirkin2018face, peng2021unified, galteri2019deep}. Recently, deep generative models are being extensively used for this task, obtaining astonishing results~\cite{nirkin2019fsgan, natsume2018fsnet, li2020advancing, peng2021unified}. Because the whole face is swapped from a source image to a target one, these methods commonly rely on the assumption that the swapped face should be recognized as the source identity. Face swapping methods based on deep networks are commonly referred to as \textit{deepFakes}. For a comprehensive survey, the reader can refer to~\cite{tolosana2020deepfakes}. 


Our proposed method extends the above in as much as we aim at swapping face parts in totally unconstrained conditions. We claim this scenario is more challenging since no assumptions on the swapped face can be made. This demands for a solution that is independent from the swapped face identity.

\begin{figure*}[!t]
\centering
\includegraphics[width=0.8\linewidth]{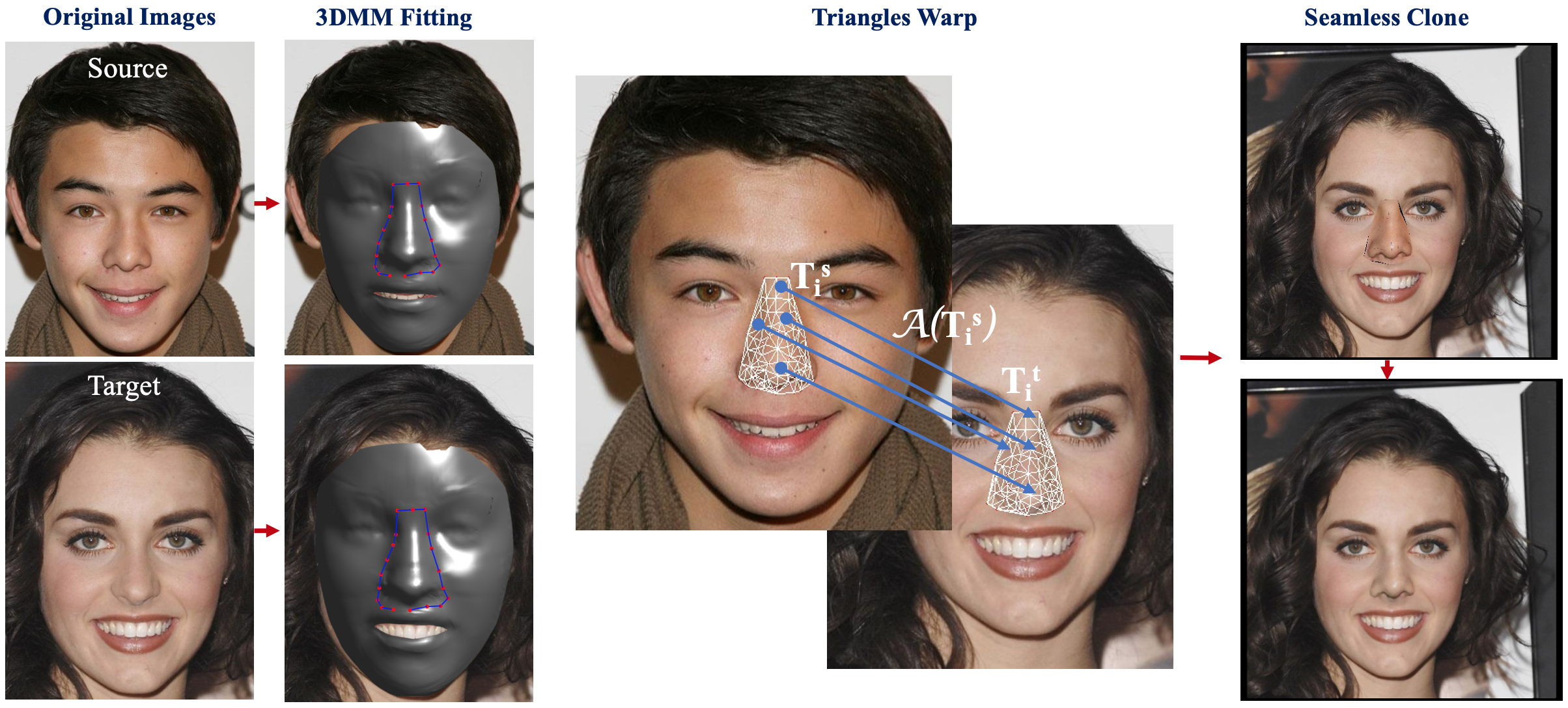}
\caption{Workflow of our proposed face parts swapping method. First, we fit a 3DMM to the face images (Sect.~\ref{subsec:3dmm}); then, we select the region of interest (Sect.~\ref{subsec:parsing}) and swap each triangular patch within the region (Sect.~\ref{subsec:triangle}). Finally, we apply Seamless Clone to merge the swapped part.}
\label{fig:pipeline}
\end{figure*}

\section{Proposed Method}\label{sec:proposed}
The proposed approach consists of three main steps: \textit{(i)} 3DMM fitting to both source and target faces; \textit{(ii)} transfer of the region of interest from source to target; \textit{(iii)} apply seamless clone to merge the transferred face region in the target image. The overall workflow is shown in Figure~\ref{fig:pipeline}. 

\subsection{3DMM Fitting}\label{subsec:3dmm}
The first step of the method involves fitting the 3DMM to the face images. This is achieved by using the \textit{Dictionary Learning} (DL) 3DMM proposed in~\cite{ferrari2015dictionary}. This particular model can accurately fit facial expressions, and has a good trade-off between fitting accuracy and computational time. The DL-3DMM is composed of an average 3D face $\mathbf{m} \in \mathbb{R}^{m \times 3}$, a dictionary of $k$ deformation components $\mathbf{D} \in \mathbb{R}^{k \times m \times 3}$, and a set of weights $\mathbf{w} \in \mathbb{R}^{k}$ to regularize the fitting. To fit the DL-3DMM to a face image, we apply the procedure illustrated in~\cite{ferrari2015dictionary}. In the following, we report some details of the process to make the description self-contained. 

We first detect 68 facial landmarks $\mathbf{l} \in \mathbb{R}^{68 \times 2}$ on the 2D images using the method proposed by Bulat~\etal~\cite{bulat2017far}. A set of 3D landmarks is extracted from the average model by indexing into the mesh, \ie, $\mathbf{L} \in \mathbb{R}^{68 \times 3} = \mathbf{m}(\mathbf{I}_L)$, being $\mathbf{I}_L \in \mathbb{N}^{68}$ the indices of the mesh vertices that correspond to the landmarks. The 2D-3D landmarks correspondence allows us to estimate the projection of the 3D landmarks $\mathbf{L}$ on the image. To this aim, we assume an orthographic camera model, which defines the relation between $\mathbf{L}$ and $\mathbf{l}$ to be:
\begin{equation}
\label{eq:simTrans}
\mathbf{l} =  \mathbf{A}\cdot\mathbf{L} + \mathbf{t} \; ,
\end{equation}

\noindent
where $\mathbf{A} \in \mathbb{R}^{2 \times 3}$ contains the affine camera parameters, and $\mathbf{t} \in \mathbb{R}^{2 \times 68}$ is the translation on the image. We first estimate the matrix $\mathbf{A}$ by solving the following least squares problem: 
\begin{equation}
\underset{\mathbf{A}}{\arg\min} \left \| \mathbf{l} - \mathbf{A} \cdot \mathbf{L} \right \|_{2}^{2}\; ,
\end{equation}

\noindent
for which the solution is given by $\mathbf{A} = \mathbf{l} \cdot \mathbf{L}^{+}$, being $\mathbf{L}^{+}$ the pseudo-inverse matrix of $\mathbf{L}$. Then, the 2D translation can be estimated as $\mathbf{t} = \mathbf{l}-\mathbf{A}\cdot\mathbf{L}$. The estimated projection $\mathbf{P} = [\mathbf{A}, \mathbf{t}]$ can be used to map the 3D landmarks onto the image with~\eqref{eq:simTrans}. All the vertices of the average face $\mathbf{m}$ can be projected on the image as well.

Using the dictionary $\mathbf{D}$, we find a set of deformation coefficients $\boldsymbol{\alpha} \in \mathbb{R}^{k}$ that non-rigidly deform $\mathbf{m}$, such that the Euclidean distance between the 2D landmarks $\mathbf{l}$ and those of the average face projected on the image plane $\mathbf{L}_p = \mathbf{A}\cdot\mathbf{L} + \mathbf{t}$ is minimized. This is achieved by solving the following regularized ridge-regression problem:
\begin{equation}
\underset{\mathbf{\alpha}}{\arg\min} \left \| \mathbf{l} - \sum_{i=1}^{k}D_i(\mathbf{I}_L)\mathbf{P}_l\alpha_i \right \|_{2}^{2} + \lambda \left \| \boldsymbol{\alpha}\cdot\mathbf{w}^{-1} \right \|_2 \; ,
\end{equation}

\noindent
where $\mathbf{D}_i(\mathbf{I}_L)$ indicates that the entries corresponding to the landmarks are extracted from the deformation components. The optimal coefficients $\boldsymbol{\hat{\alpha}}$ are used to deform $\mathbf{m}$ and obtain a fitted 3D face $\mathbf{S} \in \mathbb{R}^{m \times 3}$ as:
\begin{equation}
\mathbf{S} = \mathbf{m} + \sum_{i=1}^k D_i\alpha_i\; ,
\end{equation}

\noindent
Finally, we can use the estimated projection $\mathbf{P}$ to project $\mathbf{S}$ on the image. For more details on the fitting process, the reader can refer to~\cite{ferrari2015dictionary}.

At the end of this process, for both the source and target images, we obtain two deformed 3D faces $\mathbf{S}_s$ and $\mathbf{S}_t$, and their projections $\mathbf{P}_s$ and $\mathbf{P}_t$. The two projections are used to map the deformed 3D faces on the images,  indicated as $\mathbf{s}_s, \mathbf{s}_t \in \mathbb{R}^{m \times 2}$.

\subsection{Face and Image Parts Selection}\label{subsec:parsing}
In order to swap local face parts, we need to identify the related regions on the images, which is done exploiting the projected 3D faces. We manually label each part, \eg, eyes or nose, on the average 3D face $\mathbf{m}$ by collecting a set of indices $\mathbf{I}_p$ that define a polygonal region. Each region encloses a set of triangles $\mathcal{T} = \{\mathbf{T}_1, \cdots, \mathbf{T}_n\}$ defined by the mesh vertices. Given that $\mathbf{s}_s$ and $\mathbf{s}_t$ are in point-to-point correspondence, this also happens for the triangulations $\mathcal{T}_s$ and $\mathcal{T}_t$. Once projected on the image, they identify triangular image patches. 

At this stage, two problems arise that could lead to selecting incorrect image patches: \textit{(i)} self-occlusion due to head rotations; \textit{(ii)} occlusions of the face due to accessories such as hats, or others such as hair or hands. In both the cases, we do not want to select triangles related to occluded parts so to not transfer incorrect image patches. 

To solve \textit{(i)}, we exploit the estimated projections $\mathbf{P}_s$ and $\mathbf{P}_t$. We apply QR decomposition to $\mathbf{P}$ and, thanks to the orthogonality property of rotation matrices, we extract the 3D rotation $\mathbf{R} \in \mathbb{R}^{3 \times 3}$. We use the rotation $\mathbf{R}$ to rotate the deformed 3D faces $\mathbf{S}$ according to the estimated pose, and calculate the visibility of each 3D vertex from the novel viewpoint using the method proposed by Katz~\etal~\cite{katz2007direct}. We then select only the triangles whose vertices are estimated as visible in both the source and target faces. To account for other types of occlusions \textit{(ii)}, we follow the idea of~\cite{nirkin2018face}, and apply the BiSeNet semantic segmentation network~\cite{yu2018bisenet} to parse the face images, and retain only the vertices of the projected 3D faces $\mathbf{s}_s$ and $\mathbf{s}_t$ corresponding to valid face pixels. At the end, we obtain two filtered sets including only valid triangles. 

\subsection{Triangle Affine Warping}\label{subsec:triangle}
Previous 3D-based approaches~\cite{nirkin2018face, masi2014pose}, directly sample the RGB values from the source image exploiting the projected 3D face, and substitute the corresponding pixels in the target image. An advantage is that it is possible to directly compensate for different head poses. On the other hand, a limitation is that it is not possible to ensure a 1:1 mapping between 3D points and pixels. Depending on the resolution of the image and the 3D model (in terms of number of vertices), it can happen either that some pixels are not sampled, or multiple vertices fall on the same pixel~\cite{ferrari2016effective}. This would eventually lead to artifacts in the swapped image. To address this issue, we directly transfer the triangular patches from the source image to the target. In order to do so, we estimate an affine transformation $\mathcal{A}: \mathbb{R}^2 \to \mathbb{R}^2$ between each triangle pair $(\mathbf{T}^s_{i}, \mathbf{T}^t_{i})$. We estimate $\mathcal{A}(\mathbf{T}^s_{i})$ using the three projected vertices of the triangle, and then warp the triangle so that the image patch can be transferred to the target image. 
Finally, to realistically merge the swapped patches into the target image and account for minor artifacts, we apply Seamless Cloning~\cite{perez2003poisson} using the OpenCV implementation.

\section{Experimental Results}\label{sec:experimental}
In the evaluation, we assess the impact that swapping face parts has on face recognition based on deep networks. However, establishing a precise recognition or verification protocol, as that proposed in~\cite{nirkin2018face}, is difficult. In fact, assigning a particular identity to the partially-swapped faces is arbitrary. Thus, we defined the following experimental setup: we considered the pre-trained VggFace model~\cite{parkhi2015deep}, which was trained on 2,622 identities from a dataset that we refer to as VggDataset. From this dataset, we selected 50 random identities, 25 males and 25 females, from which we chose 4 images each. We did not put effort in choosing the identities, except that we excluded images with extremely large pose variations to avoid generation artifacts that could impair the results. 
We then performed both full and partial face swaps either across the same (\textit{intra-subject}) or different identities (\textit{inter-subject}). The goal is to collect the classifier output and evaluate how it changes when face parts are swapped.


\begin{figure}[!t]
\centering
\includegraphics[width=0.9\linewidth]{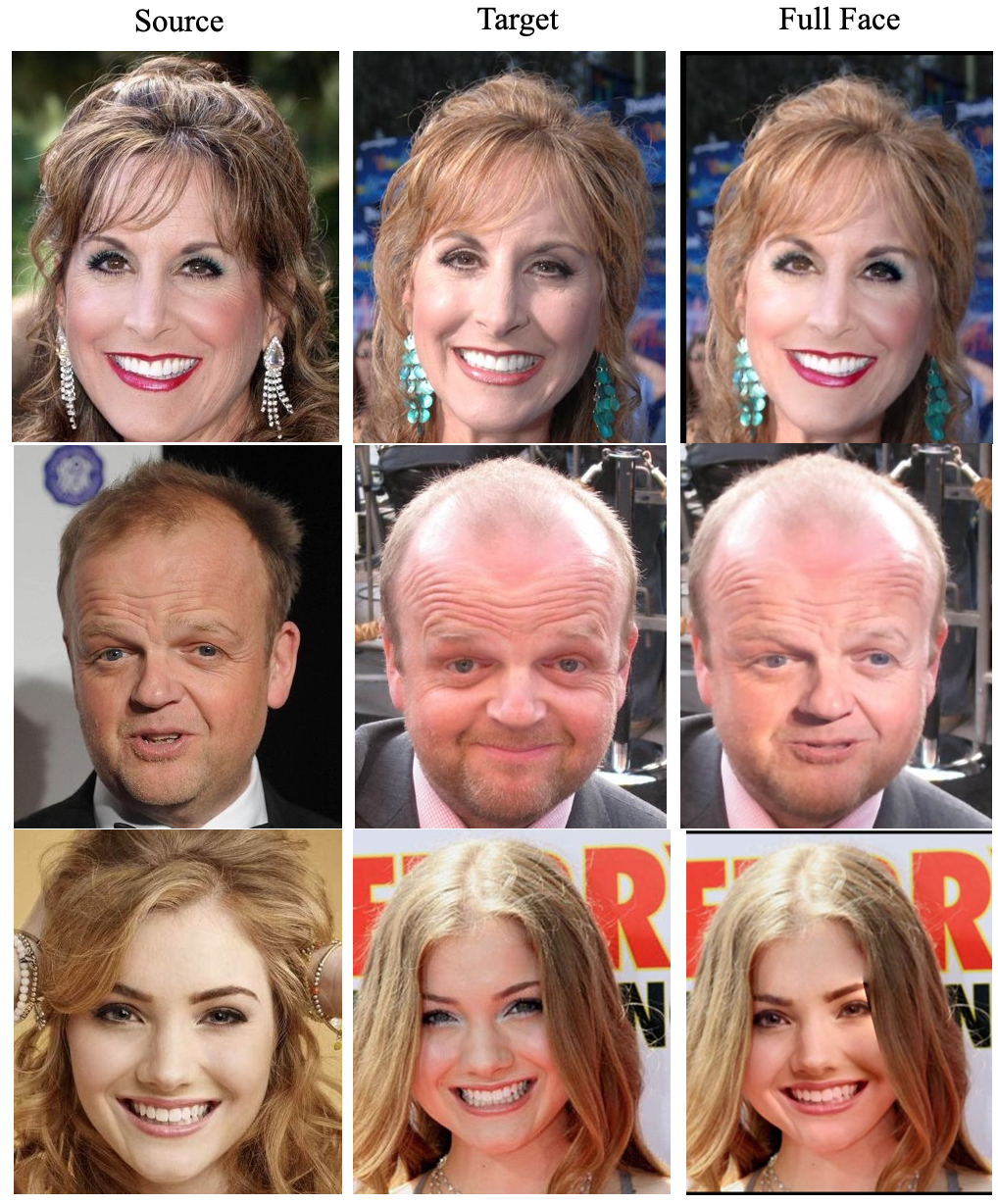}
\caption{Examples of intra-subject full face swaps.}
\label{fig:intra-swap_ex}
\end{figure}

\begin{figure*}[!t]
\centering
\includegraphics[width=0.95\linewidth]{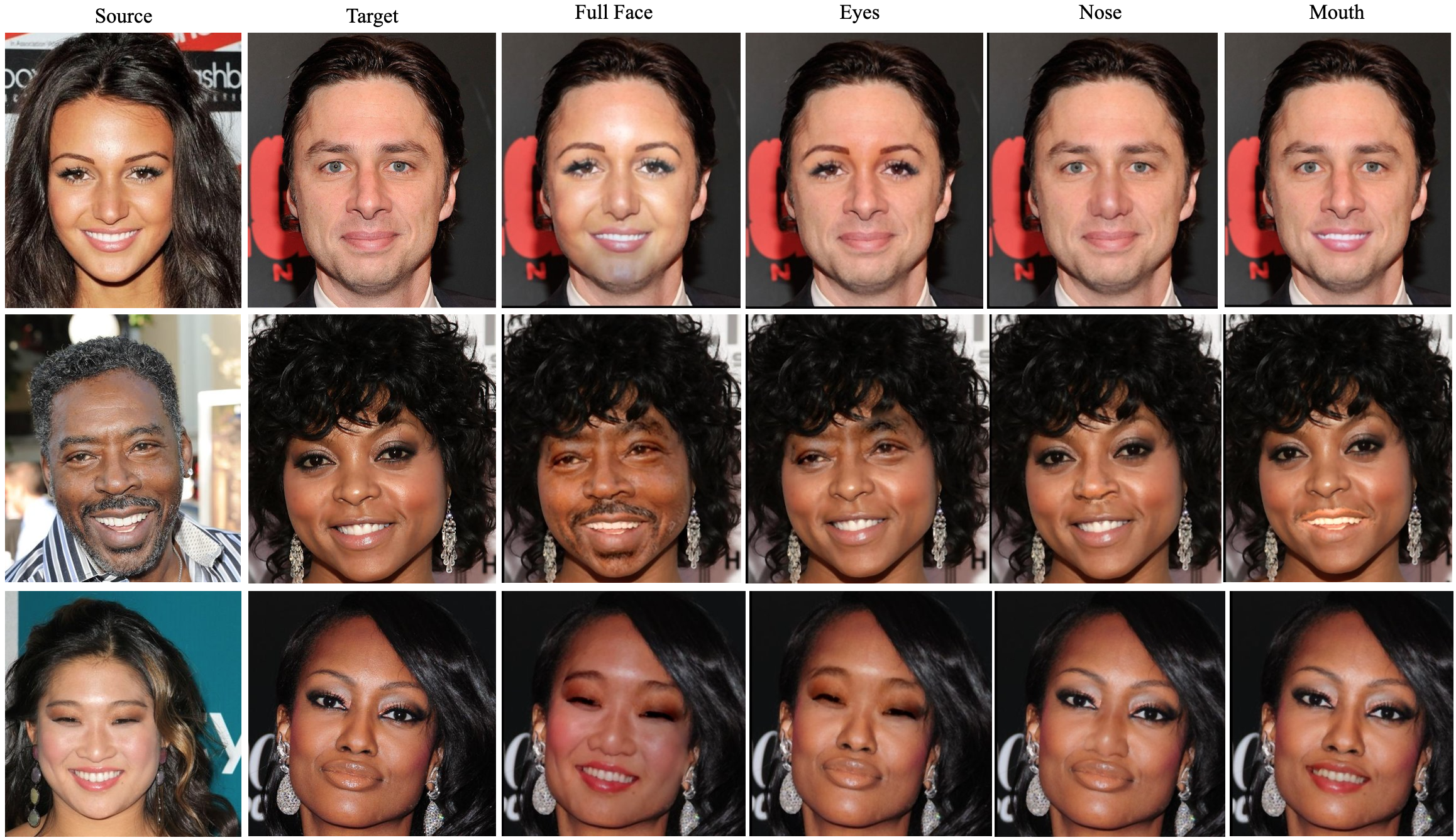}
\caption{Examples of inter-subject face swaps. For faces with very different contexts (top and middle row), swapping the full face does not necessarily imply the source identity is easily recognizable. Also, we observe that swapping the nose is barely perceptible, while the eyes severely impact the identity perception.}
\label{fig:swap_ex}
\end{figure*}

\subsection{Intra-subject swap}\label{subsec:exp-intrasub}
We performed full face swaps as well as partial swaps of mouth, nose, and eyes between images of the same identity for all the selected 50 subjects, and compared the classification accuracy. In all the cases, we assume the swapped identity must be that of the source image, given that both source and target are the same subject. Results are reported in Table~\ref{tab:intra-sub}. Performing a full face swap has a negligible impact on the classification results, highlighting that the swapping process is accurate. Some qualitative examples are shown in Figure~\ref{fig:intra-swap_ex}. The same holds for the partial swaps, even though a larger accuracy loss (around 2\% - 4\%) is observed. This is due to the difficulty of correctly swapping and merging local face regions coming from, possibly very different, images. In fact, other then adapting to different expressions, the capturing conditions, \eg, illumination, pose or scale, can change significantly.
Finally, we observe the worst result is related to the nose region, which is much more sensitive to pose variations and self-occlusions since it comes out of the face plane.


\begin{table}[!t]
\renewcommand{\arraystretch}{1.3}
\caption{Classification results for intra-subject swapping.}
\label{tab:intra-sub}
\centering
\begin{tabular}{|c|c|c|c|c|}
\hline
Original & Full face & Eyes & Nose & Mouth\\
\hline
96.42\% & 96.15\% & 94.54\% & 92.72\% & 94.67\%\\
\hline
\end{tabular}
\end{table}

\subsection{Inter-subject swap}\label{subsec:exp-intersub}
In this experiment, full face swaps, as well as partial swaps of mouth, nose, and eyes were performed between images of different identities. Considering all the possible combinations is prohibitive; so, for each identity we chose a random target subject from the remaining. In particular, we performed a balanced number of swaps selecting pairs of subjects of same or different genders. Results are reported in Table~\ref{tab:inter-sub-classif}, separating between results obtained on same gender pairs (first row), opposite gender pairs (middle row), and the overall pairs (bottom row). Here, we also report results separately for the source and target identities. In fact, when performing full or partial swaps from a source face of a subject $A$ to a target face of a subject $B$, the identity of the swapped face image could either be $A$, $B$ or neither of them. Note that the background and external facial features, \eg, hair, are those of the target subject $B$. Our goal is to quantify how frequently the swapped face is recognized as identity $A$ or $B$. They don't sum up to 100\% since the classifier could predict a different identity.

From Table~\ref{tab:inter-sub-classif}, we can observe that the recognition accuracy of a full face swap is low: only in 37.8\% of the cases, the swapped face is recognized by the network. This prompts the consideration that, consistently with the findings of~\cite{axelrod2010external, sinha1996think}, the identity information is not necessarily carried by the face only, and that external cues are also important for recognition. Clearly, this depends on the prior similarity of the two subjects. First two rows of Table~\ref{tab:inter-sub-classif} indeed show that if the swap is performed on male-male or female-female pairs, the recognition increases. Examples are shown in Figure~\ref{fig:swap_ex}: for the top and middle row examples, the full face swap is neither recognized as the source nor the target identity, while the bottom row example is recognized as the source identity, given its context similarity with the target. 

\begin{table}[!t]
\caption{Classification results for inter-subject swapping.}
\label{tab:inter-sub-classif}
\centering
\begin{adjustbox}{width=\columnwidth,center}
\begin{tabular}{|c|c|c|c|c|c|c|}
\hline
& & Original & Full face & Eyes & Nose & Mouth\\
\hline
\multirow{2}{*}{Same} & Source & -       & 51.1\% & 2.3\% & 0\% & 0\% \\
                      & Target & 99.2\%  & 6.6\%  & 37.7\% & 86.6\% & 87.3\% \\
\hline
\multirow{2}{*}{Diff} & Source & -      & 21.6\% & 0\% & 0\% & 0\%\\
                     & Target & 94.4\%  & 0\%    & 24.3\% & 97.3\% & 96.5\% \\
\hline
\hline
\multirow{2}{*}{All} &Source & -       & 37.80\% & 1.12 \%     & 0\%     & 0\%    \\
& Target & 96.15\% & 3.65\%  & 32.92\%     & 91.64\% & 92.45\%\\
\hline
\end{tabular}
\end{adjustbox}
\end{table}

\begin{table}[!t]
\caption{Rank@1 recognition rates for inter-subject swapping.}
\label{tab:inter-sub-recog}
\centering
\begin{adjustbox}{width=\columnwidth,center}
\begin{tabular}{|c|c|c|c|c|c|c|}
\hline
& & Original & Full face & Eyes & Nose & Mouth\\
\hline
\multirow{2}{*}{Inceptionv1~\cite{szegedy2016rethinking}} & Source & - & 86.6\% & 26.8\% & 2.4\%  &  1.2\% \\
& Target & 90.2\% & 9.7\% & 60.9\% & 86.7\% & 86.5\%\\
\hline
\multirow{2}{*}{ResNet50~\cite{he2016deep}} & Source & - & 70.7\% & 13.4\% & 0\%    &   0\% \\
                          & Target & 88.7\% & 13.4\% & 60.9\% & 80.5\$ & 81.7\%\\
\hline
\multirow{2}{*}{SENet~\cite{hu2018squeeze}} & Source & -      & 65.8\% & 12.2\% & 0\%    &   0\% \\
                       & Target & 85.4\% & 20.7\% & 63.4\% & 75.6\$ & 78.5\%\\
\hline
\end{tabular}
\end{adjustbox}
\end{table}

A second interesting outcome is related to the swap of local parts. When swapping either the nose or the mouth, the recognition is not really compromised, and the target identity $B$ (the original one) is almost always correctly recognized. Instead, when swapping the eyes region (including eyebrows), the accuracy drops dramatically. In a handful of cases, we can even make the classifier predict the source identity $A$ only by swapping its eyes. In 2006, Sinha~\etal~\cite{sinha2006face} already investigated this particular effect in human perception, pointing to the conclusion that eyes and eyebrows are the most prominent features used by humans for recognizing faces; here, we quantitatively demonstrate that this effect also occurs in machine vision systems. 
We note that this behavior is consistent also for the subsets of swaps performed on same or different gender pairs, even though swapping nose or mouth on same gender pairs make the classifier wrong more often. Given the similarity of external cues, the network is likely to be more sensitive to internal face features.

\begin{figure}[!t]
\centering
\includegraphics[width=0.9\linewidth]{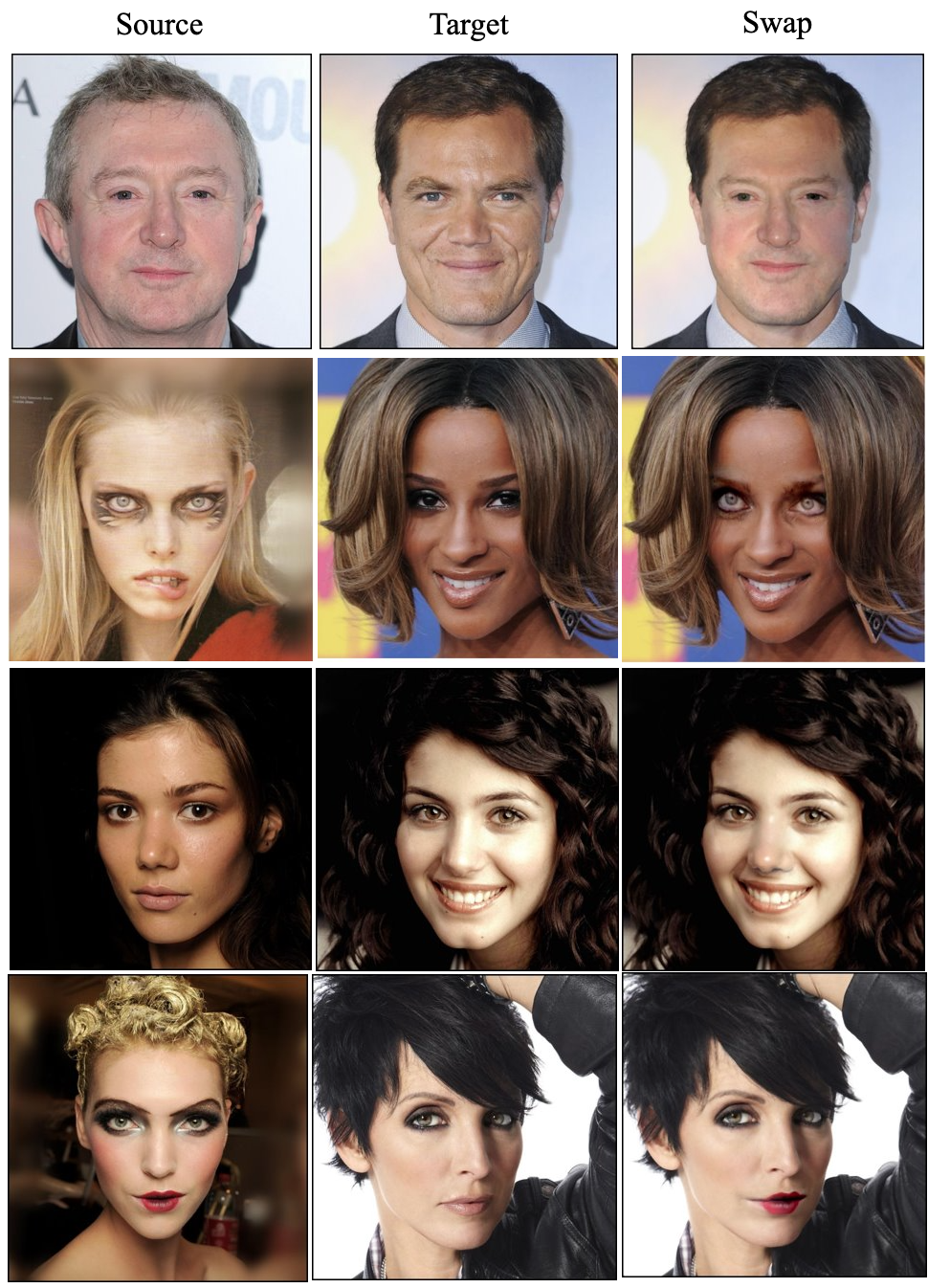}
\caption{Examples of inter-subject face swaps on the CelebMaskHQ dataset. From top to bottom, we show full face, eyes, nose and mouth swaps.}
\label{fig:celebA_ex}
\end{figure}

\begin{figure}[!t]
\centering
\includegraphics[width=0.92\linewidth]{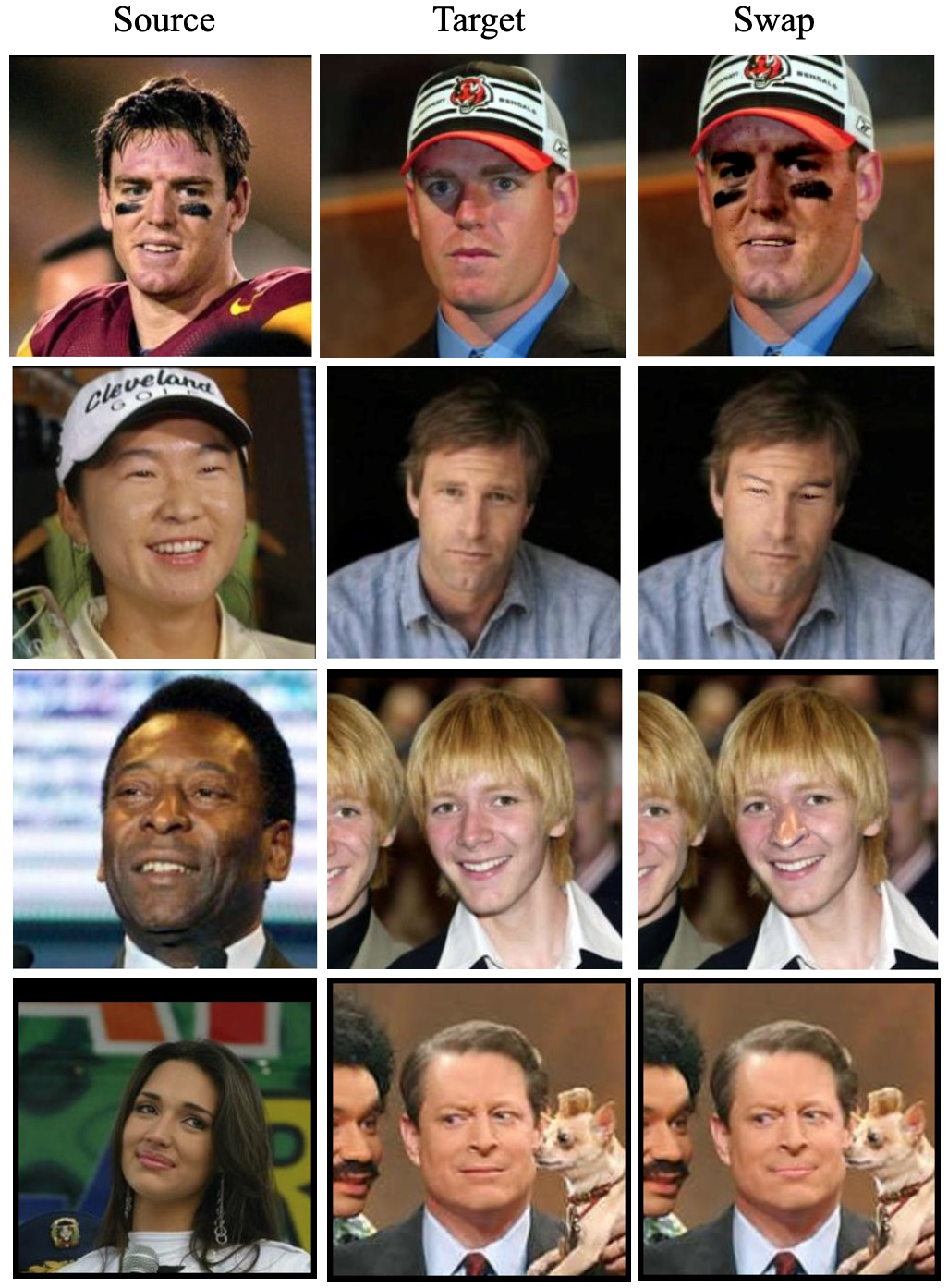}
\caption{Examples of inter-subject face swaps on the LFW dataset. From top to bottom, we show full face, eyes, nose and mouth swaps.}
\label{fig:lfw_ex}
\end{figure}

\noindent
\textbf{Identification protocol:} In order to verify the above results in a more general scenario, we devised a standard Gallery-vs-Probe identification protocol. We chose one image for each of the 50 subjects to compose the gallery, and used all the swapped images as probe (query) set. Note that the images are the same as those used in Sect.~\ref{subsec:exp-intersub}. We used three different architectures, namely, InceptionV1~\cite{szegedy2016rethinking}, pre-trained on CASIA WebFace, ResNet50~\cite{he2016deep} and SENet~\cite{hu2018squeeze}, both pre-trained on VggFace2~\cite{cao2018vggface2}, to extract face descriptors from all the images. Recognition is then performed following the standard practice: for each probe face, we compute the cosine similarity against all the samples in the gallery, and pick the most similar identity. Rank@1 results are reported in Table~\ref{tab:inter-sub-recog}.

From Table~\ref{tab:inter-sub-recog}, we can observe the behavior follows a similar trend with respect to that reported in Table~\ref{tab:inter-sub-classif}, even if the accuracy is generally lower. For all the three networks, we observe performing full face swaps make the descriptors match with that of the source identity $A$ most of the times, being fooled by the manipulation significantly more often with respect to VggFace. A reason is that the tested networks were trained using tighter bounding boxes, with less contextual information. Performing nose and mouth swaps instead, similar to Table~\ref{tab:inter-sub-classif}, does not have a major impact on the results, and the target identity $B$ is correctly matched in the majority of cases. Again, we observe instead a way more significant change in the identification when swapping the eyes. The identification rate of the target drops of around 25\%/30\%, while the source identity (from which only the eyes have been taken) increases significantly. These results strongly support our previous claim, and other literature outcomes related to human face perception.

\noindent
\textbf{Tests on CelebMaskHQ and LFW:} Finally, we qualitatively evaluated our swapping method also on the CelebMaskHQ~\cite{lee2020maskgan} and Labeled Faces in the Wild (LFW)~\cite{huang2008labeled} datasets. This is intended to demonstrate that our strategy allows applying the method on images of arbitrary resolution. Figure~\ref{fig:celebA_ex} and~\ref{fig:lfw_ex} report some examples.

\section{Discussion and Limitations}\label{sec:limitations}
In this paper, we proposed a method for swapping local face parts, with the goal of analyzing how that impacts on face recognition. We extended previous studies on face perception at a finer level, proposing some preliminary investigations on previously unexplored aspects of face recognition based on deep learning. Furthermore, the ability of swapping local face parts can be useful for many tasks, from improving robustness to disguised faces or manipulation attacks, to entertainment or medical applications. At the current stage, the proposed method has yet some limitations. First, even though we can handle natural pose variations, large pose differences induce artifacts. Also, using seamless clone to merge the swapped part is effective in most of the cases, but does not work well in extreme scenarios, \eg, mixing color and gray-scale images. Finally, swapping parts from low- to high-resolution images results in blurry and visible swaps. For future developments, we aim at investigating the use of deep generative models to apply local face manipulations.

{\small
	\bibliographystyle{./IEEEtran}
	\bibliography{bibliography}
}

\end{document}